\documentclass{article}
\usepackage{spconf,amsmath,graphicx} 

\usepackage{enumitem}
\setlist{nosep,leftmargin=14pt}

\usepackage{mwe} 
\usepackage{booktabs} 
\usepackage{multirow}
\usepackage{tabularx}
\usepackage{adjustbox}
\usepackage{array}
\usepackage{gensymb}
\usepackage{anyfontsize}

\usepackage{amssymb}
\usepackage{pifont}
\usepackage{hyperref}

\newcommand{\cmark}{\ding{51}}%
\usepackage{adjustbox}

\title{Improving Human Sperm Head Morphology Classification with Unsupervised Anatomical Feature Distillation}
\name{Yejia Zhang\thanks{Corresponding author contact: yzhang46@nd.edu.}$^\dagger$
     \quad Jingjing Zhang$^\star$
     \quad Xiaomin Zha$^\star$ 
     \quad Yiru Zhou$^\star$ 
     \quad Yunxia Cao$^\star$ 
     \quad Danny Z. Chen$^\dagger$ }
\address{ $^\dagger$University of Notre Dame, Department of Computer Science and Engineering, Notre Dame, IN, USA \\
          $^\star$First Affiliated Hospital of Anhui Medical University, Obstetrics and Gynecology,
                        Hefei, Anhui, China}
\begin{document}
%

\maketitle

\begin{abstract}
With rising male infertility, sperm head morphology classification becomes critical for accurate and timely clinical diagnosis. 
Recent deep learning (DL) morphology analysis methods achieve promising benchmark results, but leave performance and robustness on the table by relying 
on limited and possibly noisy class labels. 
To address this, we introduce a new DL training framework that leverages anatomical and image priors from human sperm microscopy crops to extract useful features without additional labeling cost.
Our core idea is to distill sperm head information with reliably-generated pseudo-masks and unsupervised spatial prediction tasks.
The predicted foreground masks from this distillation step are then leveraged to regularize and reduce image and label noise in the tuning stage. 
We evaluate our new approach on two public sperm datasets and achieve state-of-the-art performances (e.g. 
65.9\% SCIAN accuracy and {96.5\% HuSHeM accuracy}).

\end{abstract}

\vspace{-1mm}
\section{Introduction}
\label{sec:intro}
\vspace{-1mm}

Sperm counts have halved in Western nations since the 1970s \cite{Levine2017TemporalTI}, and use of Assisted Reproductive Technology (ART) has increased to account for over 2\% of American births annually \cite{ART2019}.
Morphology analysis, a central step in infertility diagnosis and ART planning, involves classifying sperm heads into qualitative shape categories \cite{WHOManual}, where abnormality predicts poor genetic fitness.
This process is both time-consuming and subjective since it requires manual deliberation on individual sperm cells in microscopy images.
These challenges, along with high annotation costs, result in small public datasets (e.g., HuSHeM \cite{Shaker2018Hushem} has 216 131$\times$131 crops; SCIAN \cite{Chang2017GoldstandardFC} has 1132 35$\times$35 crops) with noisy class labels.


\begin{figure}[htb]
    \begin{minipage}[b]{1.0\linewidth}
      \centering
       \centerline{\includegraphics{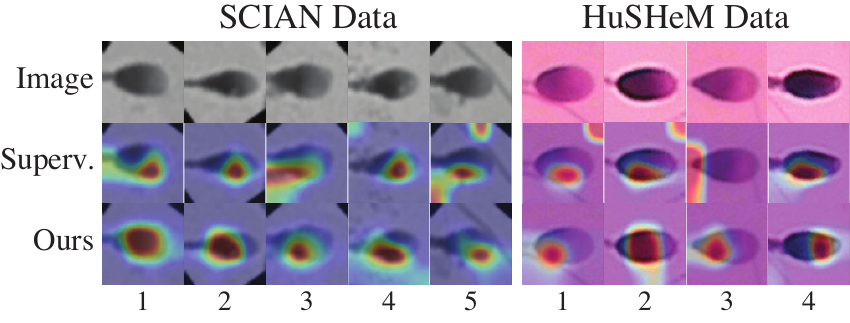}}
    \end{minipage}
    %
    \vspace{-7mm}
    \caption{Stained human sperm head samples and CAM \cite{Selvaraju2019GradCAMVE} comparisons of two morphology datasets. Row 1: 64$\times$64 center-cropped, right-aligned test images. Rows 2 and 3: CAMs with supervised training and training with our framework, respectively. The 5 indices are for the classes of normal, tapered, pyriform, amorphous, and small, respectively.}
    \label{fig:1}
    \vspace{-3mm}
\end{figure}

Two main approaches were used to address small datasets in sperm morphology analysis: 1) manually design shape features; 2) use transfer learning with neural networks (NNs).
Previous work with the first approach applied traditional machine learning (ML) methods without relying on big data.
Notably, Chang et al.~applied a cascade ensemble of support vector machines (CE-SVM \cite{Chang2017AutomaticCO}) to classify sperm heads using various shape descriptors (e.g., area, eccentricity, Zernike moments); Shaker et al.~\cite{Shaker2017ADL} presented a dictionary learning method (APDL) based on prototype matching of image patches. These methods yielded reasonable results, but methods using approach two \cite{Riordon2019DeepLF,Iqbal2020DeepLM,Liu2021AutomaticMA} generally attained superior performance with ImageNet pretrained models.


\begin{figure*}[htb]
    \centering
    \begin{minipage}[b]{0.95\linewidth}
      \centering
      \centerline{\includegraphics[width=15cm]{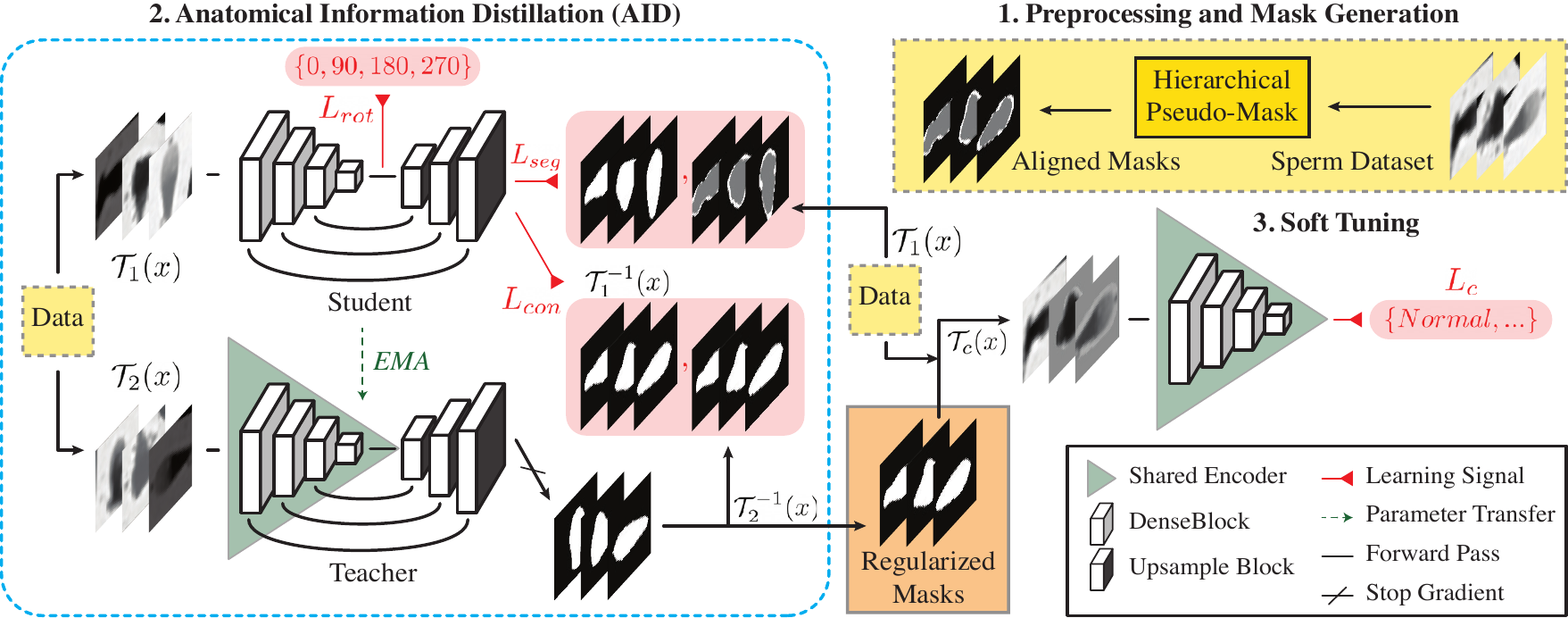}}
    \end{minipage}
    \vspace{-3mm}
    \caption{An overview of our three-stage framework. 
    EMA is for exponential moving average.
    Dashed-yellow boxes represent data sources containing HPM-generated masks while the solid-orange box has masks predicted by the teacher model in AID.}
    \label{fig:2}
    \vspace{-3mm}
\end{figure*}

Despite performance gains from NNs, effective and robust features are difficult to learn with noisy labels and small data (both 
in terms of crop size and label-set cardinality).
In Fig.~\ref{fig:1}, we notice hints of brittleness where salience is concentrated on limited discriminative foreground regions and some unrelated background patches.
Regularized spatial pretraining is shown to help steer salience towards the foreground (Fig.~\ref{fig:1} row 3), while simultaneously encoding useful shape, texture, and general sperm features.
Given our ability to cheaply generate reliable, information-dense foreground masks with useful priors, we elect to pursue a transfer learning approach.



To the best of our knowledge, no other work has used pseudo-masks with spatial prediction tasks to facilitate classification. 
The converse was introduced \cite{Zhang2019ASO}, where hints from classification were used to improve localization.
Relevant to the spatial pretraining in this work, a student-teacher setup \cite{Li2021TransformationConsistentSM} was proposed to regularize segmentation predictions in a semi-supervised setting.
We use a similar setup except for unsupervised pretraining with no pure ground truths available.
Additionally, rotation prediction \cite{Gidaris2018UnsupervisedRL} was used as an unsupervised pretext task for natural scene images.
We use this method for a similar purpose, except we are operating under stricter constraints since the upright orientation assumption does not hold in our situation.
As for sperm mask generation, other traditional vision methods \cite{Shaker2016AutomaticDA,Chang2014GoldstandardAI} have been proposed, but they either did not apply to gray images or were not calibrated to handle low quality, noisy images (e.g., in SCIAN).

In this work, we leverage anatomical and image priors of sperm images to improve our model's salience behavior and 
performance by proposing a new three-stage deep learning framework. (1) Reliable sperm head segmentations are generated in a hierarchical manner using the size, shape, contrast, location, and data priors.
(2) Anatomical Information Distillation (AID) uses the masks to distill useful foreground information 
via two unsupervised spatial prediction tasks.
(3) Soft-tuning
regularizes noisy class labels by implementing a weighted soft loss and uses the refined AID-generated masks for background denoising.

Our main contributions are as follows.
\begin{itemize}
    \item We propose a new three-stage sperm morphology classification framework that uses generated pseudo-masks to guide the learning of useful sperm features via two unsupervised spatial prediction tasks. The learned parameters and refined mask predictions are transferred to the soft-tuning stage where label smoothing and dynamic background subtraction help alleviate label and image noise.
    \item We introduce a hierarchical sperm head segmentation method that leverages anatomical and image priors to reliably filter imaging artifacts, remove sperm tails, and discard the sperm mid-pieces. 
    \item We evaluate our new framework on two public 
    sperm morphology datasets (SCIAN \cite{Chang2017GoldstandardFC} and HuSHeM \cite{Shaker2018Hushem}) and achieve state-of-the-art performances on both.
\end{itemize}

\section{Method}
\vspace{-2mm}
\label{sec:methods}

In this section, we detail our methodology on each of the three main framework components (shown in Fig.~\ref{fig:2}).

\vspace{-3mm}
\subsection{Hierarchical pseudo-mask generation (HPM)}
\vspace{-4mm}
\begin{figure}[htb]
    \centering
    \begin{minipage}[b]{0.6\linewidth}
      \centering
      \centerline{\includegraphics[width=5.85cm]{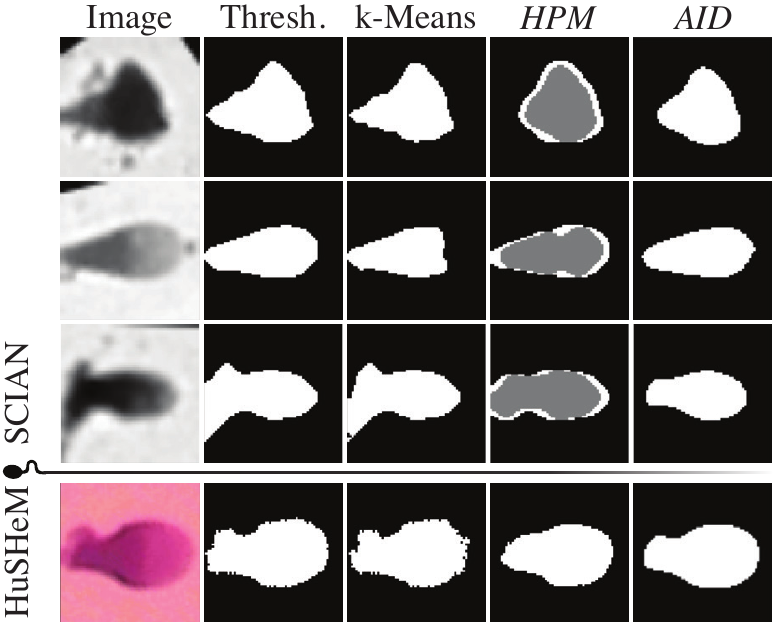}}
    \end{minipage}
    \vspace*{-2mm}
    \caption{Qualitative pseudo-mask results: comparing 
    our methods with traditional mask generation methods. The \emph{HPM} column shows results from Hierarchical Pseudo-Masks, where $h=2$ is used for SCIAN (gray is the most confident layer) and $h=1$ for cleaner HuSHeM data. \emph{AID} shows masks predicted by Anatomical Information Distillation.}
    \label{fig:3}
\end{figure}
\vspace{-2mm}

HPM has two key objectives: create accurate sperm head masks, and use the computed head orientation to align all sperm to point right (see \cite{Liu2021AutomaticMA} for the importance of this step).
A sperm cell consists of three main components: 1) the head (or foreground), 2) a mid-section that often has similar texture and intensity as the head, and 3) a tail which is a thin organelle connecting to the mid-section.
The most challenging aspect of head segmentation lies in the separation of the mid-section from the head (in rows 1 \& 3 of Fig.~\ref{fig:3}, one can see over-segmentation of the mid-section by thresholding, due to its connectivity and similar appearance).
To produce reliable masks, we leverage both anatomical priors (i.e., head size, tapered nuclear shape, darker nuclear intensity, rough elliptical outline) and image priors (i.e., relatively centered foreground, one instance per image, and contrast from dying agents).
We present a high-level flow of our proposed method below.

\noindent
(1) For each image $I$, a coarse foreground mask, $\tilde{m}_1$, is created by applying non-local means denoising, followed by thresholding. $\tilde{m}_1$ is used for intensity-range rescaling of the foreground and whitening of the background, to produce $I_1$.

\noindent
(2) Extract the darkest segment of $I_1$ with k-means clustering ($k=3$) to obtain a tentative nuclear mask. The tail and mid-section pieces are removed by distance thresholding to yield a reliable nuclear mask, $\tilde{n}_1$. An ellipse is fitted to $\tilde{n}_1$ and the thicker end of $\tilde{n}_1$ is chosen as the head direction. Finally, $\tilde{m}_1$, $I_1$, and $\tilde{n}_1$ are rotated to face right, outputting $\tilde{m}_2$, $I_2$, and $\tilde{n}_2$.

\noindent
(3) Fuse $\tilde{n}_2$ with a reliable acrosome mask (the top part of the head). Since we already computed the orientation, we simply split $\tilde{m}_2$ in half and use the right part as a coarse acrosome mask, $\tilde{m}^r_2$. We erode $\tilde{m}^r_2$ until it is within a threshold of $\tilde{n}_2$'s height, and fuse them to produce $\tilde{M}_0$, where $0$ indicates the lowest hierarchy index or the most confident mask.

\noindent
(4) Obtain a final hierarchy of concentric masks, $\tilde{M}$. For a given $h\in\mathbb{N}$, we obtain $\tilde{M}_i$ for $i \in \{0, ..., h-1\}$, a mask layer, by dilating $\tilde{M}_0$ $i$ times with all extents bounded by $\tilde{m}_1$.  

\vspace{-2mm}
\subsection{Anatomical information distillation (AID)}
\vspace{-1mm}

We utilize a student-teacher setup \cite{Tarvainen2017MeanTA} with two fully convolution networks where the teacher parameters are updated as an exponential moving average (EMA) of the student's parameters after each iteration.
The student model consists of a backbone encoder $f_s$ with an upsampling decoder $g_s$.
The teacher network shares an identical architecture, denoted as $f_t$ and $g_t$ (see AID in Fig.~\ref{fig:2}).
$\mathcal{T}_1$ and $\mathcal{T}_2$ are stochastic image transformation functions that use the same set of augmentations. 
AID first performs fine distillation with segmentation and consistency regularization, and then
conducts coarse distillation with rotation prediction.


\vspace{0.025in}
\noindent
\textbf{Fine distillation.}  
For an image $I$ in a dataset $\mathcal{D}$ (e.g. oriented crops from HPM), the predicted segmentation masks from the student and teacher networks are $\hat{y}_s = g_s(f_s(\mathcal{T}_1(I)))$ and $\hat{y}_t = g_t(f_t(\mathcal{T}_2(I)))$, respectively. 
Partial cross-entropy is used for the segmentation loss where $\tilde{M}_0$ is selected as the foreground ground truth, $\lnot\tilde{M}$ is the binary background ground truth ($\lnot$ indicates the pixel-wise negation operator that turns nonzero foreground pixels into background zeros and vice-versa), and all other non-confident pixels, $\tilde{M}_{n}$ where $n>0$, are ignored during pretraining.
Note that ignored pixels only exist if $h>1$ hierarchies are generated since $h=1$ yields one foreground mask.

Furthermore, to prevent overfitting, we add a consistency constraint where two transformed versions of the same image are encouraged to have identical predictions after mask alignment.
The two fine losses are defined as:

\vspace*{-4mm}
\begin{equation}
    \mathcal{L}_{seg} = \frac{-1}{|I|} \sum_{i \in I} (\lnot\tilde{M}_{n}^i) (\lnot\tilde{M}^i \cdot \log(1 - \hat{y}_{s}^i) + \tilde{M}_{0}^i \cdot \log(\hat{y}_{s}^i)) \\
\end{equation}
\vspace*{-3mm}
\begin{equation}
    \mathcal{L}_{con} = ||\mathcal{T}_1^{-1}(\hat{y}_{s}) - \mathcal{T}_2^{-1}(\hat{y}_{t})||^2_2,
\end{equation}
where $i$ is a pixel in image $I$. 
The aggregate fine distillation loss is $\alpha \mathcal{L}_{seg} + \beta \mathcal{L}_{con}$, with balancing parameters $\alpha$ and $\beta$.
$\hat{y}_t$ masks are also saved for use in the soft-tuning stage.

\newcommand{\ra}[1]{\renewcommand{\arraystretch}{#1}}
\setlength{\tabcolsep}{4.75pt}{
\begin{table*}[htb]
    \centering
    \ra{0.925} 
    \begin{tabular*}{1.0\linewidth}{@{}llllllllll@{}}
        \toprule
        \multirow{2}{*}[-0.15em] {\hspace{7pt} \textbf{Method}} &
        \multicolumn{4}{c}{SCIAN Dataset} & {} & \multicolumn{4}{c}{HuSHeM Dataset} \\
        \addlinespace[0pt]
        \cmidrule(lr){2-5} \cmidrule(lr){7-10}
        {} & Accuracy & Recall & Precision & F$_1$ &&
        Accuracy & Recall & Precision & F$_1$ \\
        \midrule
        CE-SVM \cite{Chang2017AutomaticCO} & 44 & 58 & - & -  && 
                        78.5 & 78.5 & 80.5 & 78.9 \\
        APDL \cite{Shaker2017ADL}   & 49 & 62 & - & - && 
                                      92.2 & 92.3 & 93.5 &  92.9 \\
        FT-VGG \cite{Riordon2019DeepLF} & 49 & 62 & 47 & 53 && 
                                      94.0 & 94.1 & 94.7 & 94.1\\
        MC-HSH \cite{Iqbal2020DeepLM} & 63 & 68 & 56 & 61 && 
                                      95.7 & 95.5 & 96.1 & 95.5 \\
        TL \cite{Liu2021AutomaticMA} & - & 62 & - & - && 
                                      96.0 & 96.1 & 96.4 & 96.0 \\
        Ours \small $\pm$Std. & \textbf{65.9} \small $\pm$ 0.68 
                            & \textbf{68.9} \small $\pm$ 0.30 
                            & \textbf{58.7} \small $\pm$ 1.20 
                            & \textbf{63.2} \small $\pm$ 0.23  
                            & 
                            & \textbf{96.5} \small $\pm$ 0.36 
                            & \textbf{96.6} \small $\pm$ 0.40 
                            & \textbf{96.8} \small $\pm$ 0.29 
                            & \textbf{96.5} \small $\pm$ 0.38 \\
        \bottomrule
    \end{tabular*}
\vspace*{-3mm}
\caption{Performance comparison of our proposed framework with state-of-the-art classification methods on the 
SCIAN (partial agreement subset) and HuSHeM datasets. Some SCIAN numbers lack decimals since they are taken directly from other works.}
\label{tab:1}
\vspace*{-3mm}
\end{table*}
}


\vspace{0.05in}
\noindent
\textbf{Coarse distillation.} 
For better adaptation to the downstream classification task, we use the rotation prediction task with four angles, $\{0, 90, 180, 270\}$, from a right-facing orientation to transition the model from a pixel-wise focus toward learning more abstract features of different anatomical parts and their respective spatial relations.
The loss is defined as:
\vspace{-1mm}
\begin{equation}
    \vspace{-1mm}
    \mathcal{L}_{rot} =  - y_{rot} \cdot \log(f_s(\mathcal{T}_1(I))_{rot}).
\end{equation}
where $y_{rot}$, $f_s(\mathcal{T}_1(I))_{rot}$, and $I$ are the target rotation index, predicted rotation index, and pre-aligned image, respectively.

\vspace{-3mm}
\subsection{Soft-tuning}

We transfer the trained teacher encoder, $f_t$, for morphology classification tuning.
We implement three key components for the purposes of regularization, noise reduction, and prediction stabilization. 
First, we use AID-predicted segmentations to mask out image backgrounds following a curriculum learning approach where we apply more aggressive masking as training progresses.
This is implemented by linearly decreasing the number of applied foreground mask dilations from 15 (entire image is foreground) to 0 (only sperm head is inputted).
    Second, we use a soft cross-entropy loss that handles label variability between annotators:
\vspace{-1mm}
\begin{equation}
    \vspace{-1mm}
    \mathcal{L}_{c} =  \lambda \cdot \log \ \hat{y}_{c1} + (1 - \lambda) \cdot \log  \ \hat{y}_{c2},
\end{equation}
where $c1$ \& $c2$ represent an image's majority \& minority classes, respectively, and
$\hat{y}_{c1}$ \& $\hat{y}_{c2}$ are the model scores for the majority \& minority classes (during consensus, the loss reduces to vanilla cross-entropy).
$\lambda \in [0,1]$ is a balancing term that represents the majority class weighting.
Finally, we introduce shape-invariant test-time augmentations (i.e., vertical flipping and rotation) to lower prediction variability.
We also make training augmentations, $\mathcal{T}_c$, milder, which empirically improves performance and stabilizes training.

\vspace{-1mm}
\section{Experiments}
\label{sec:experiments}
\vspace{-2mm}
\subsection{Datasets}
We use the only two publicly available stained sperm morphology datasets for evaluation. 
SCIAN \cite{Chang2017GoldstandardFC} (from the \textbf{Sc}ientific \textbf{I}mage \textbf{An}alysis Lab, University of Chile) contains 1132 35$\times$35 gray crops with five morphology classes (100 normal, 228 tapered, 76 pyriform, 656 amorphous, and 72 small).
The images were captured at 63X zoom after Hematoxylin \& Eosin staining.
Three experts assigned classes for each crop. The partial agreement subset we use contains only images with at least 2 out of 3 experts agreeing on a label.

HuSHeM \cite{Shaker2018Hushem} (\textbf{Hu}man \textbf{S}perm \textbf{He}ad \textbf{M}orphology) has 216 131$\times$131 RGB crops and four classes (54 normal, 53 tapered, 57 pyriform, 52 amorphous) with expert consensus.
Images were obtained at 100X magnification after Diff-Quik staining.
Fig.~\ref{fig:1} shows pre-processed HuSHeM and SCIAN samples.

\vspace{-2mm}
\newcolumntype{R}[2]{%
    >{\adjustbox{angle=#1,lap=\width-(#2)}\bgroup}%
    l%
    <{\egroup}%
}
\newcommand*\rot{\multicolumn{1}{R{45}{1em}}}
\begin{table}[htb]
\begin{tabular}{rrrrrrrcc} 
\rot{\small ImNet Pre.} &
\rot{\small Head Pred.} & 
\rot{\small Rot. Pred.}  &
\rot{\small TTA} &
\rot{\small Soft Lab.} &
\rot{\small Dyn. Mask} & &
Accuracy & 
F$_1$ \\ \hline
\cmark & 		         & 		     & 		 &  		& 		  & 		    &  \textbf{51.1}  & \textbf{53.7} \\ 
               & \cmark &  &  & &  &                                             & 50.9 & 53.6   \\ 
               &                & \cmark & & & &                                 & 48.3 & 52.4  \\  \hline
 \cmark & \cmark & \cmark &                &  & &                  & 58.6 & 57.3 \\
 \cmark & \cmark & \cmark & \cmark &  & &                  & 59.1 & 58.7 \\
 \cmark & \cmark & \cmark & \cmark & \cmark &  &   & 63.3 & 61.5\\
 \cmark & \cmark & \cmark & \cmark & \cmark & \cmark & & \textbf{65.9} & \textbf{63.2}\\
\end{tabular}
\vspace{-1mm}
\caption{Ablation study results 
on all five SCIAN folds (3 runs per fold). TTA stands for test-time augmentation.}
\label{tab:2}
\vspace{-6mm}
\end{table}

\subsection{Anatomical information distillation (AID)}
\label{exp:pretrain}

\noindent
\textbf{Implementation details.}
We follow the 
5-fold cross-validation procedure 
in \cite{Iqbal2020DeepLM,Liu2021AutomaticMA}.
Each data fold is pretrained separately to avoid information leakage from the hold-out set. 
The backbone model is an ImageNet-pretrained DenseNet-201 \cite{Huang2017DenselyCC}. 
We attach bilinear upsampling and additive fusion layers for segmentation, and a linear layer to the lowest resolution features for rotation prediction.
For augmentations ($\mathcal{T}_1$, $\mathcal{T}_2$), we resize the image to 64x64, apply 10 degree rotation, vertical flip, 10\% shift, hue/saturation jitter, gray conversion, 50\% brightness change, 20\% contrast change, $[0.6, 1]$ scaling, and $[0.6, 1.5]$ aspect ratio distortion for a final 64$\times$64 crop.
We train with batch size 128 for 2500 iterations (25 epochs) using default Adam (0.0001 learning rate), and a step-decay scheduler annealing the learning rate by 0.1 at epochs 12 and 19.
For the fine loss weights, we use $\alpha=1$ and $\beta=1$.


\vspace{1mm}
\noindent
\textbf{Results.}
We present qualitative results comparing our HPM and AID with some traditional methods in Fig.~\ref{fig:3} since we have no segmentation ground truth.
To gauge pretraining effectiveness, we present downstream performance in Table \ref{tab:1} and ablation study results in Table \ref{tab:2}.

\vspace{-2mm}
\subsection{Soft-tuning}
\vspace{-1mm}
\textbf{Implementation details.}
We maintain separate folds for fair comparison with previous work, and use the same splits as in Section \ref{exp:pretrain}.
The AID-pretrained DenseNet-201 backbone is used for tuning and a new linear layer is appended for classification.
For SCIAN [HuSHeM] augmentations, $\mathcal{T}_c$, we resize images to 64x64 [centercrop + 64x64], apply 10\degree \ [0\degree] rotation, vertical flip, 6\% [10\%] shift, 0/0 [50/30] hue/saturation jitter, [grayscale conversion], and 0\%/0\% [50\%/20\%] brightness/contrast change to obtain a final 64$\times$64 crop.
We train with batch size 512 [256] for $\sim$1500 iterations (30 epochs) with default Adam (0.00015 learning rate), and a step-decay scheduler annealing the learning rate by 0.1 at epochs 14 and 23.
For the soft cross-entropy loss, we use $\lambda=0.85$.
\vspace{-3mm}
\\~\\
\textbf{Results.}
Table \ref{tab:1} shows the comparison results of our framework and previous methods on the two datasets.
We use four metrics as the main comparison measures based on precedence in the literature.
On the challenging SCIAN dataset, we improve over \cite{Iqbal2020DeepLM} on all metrics by at least 2.2\% except on recall.
For HuSHeM, we also outperform in every metric except the increase is smaller because of diminished margin for improvement from the previous art's high scores.

\vspace{-3mm}
\subsection{Ablation study}
\vspace{-1mm}

Quantitative results for ablation study are given in Table \ref{tab:2}.
The first three rows show that when we compare different pretraining paradigms individually, ImageNet-pretrained leads.
However, from row 4, we see that aggregated pretraining (AID) begins to outperform.
Finally, in the bottom three rows, one can see the contributions from each of our proposed soft-tuning components to improve training.

\vspace{-2mm}
\section{Conclusions}
\label{sec:end}
\vspace{-1mm}
We proposed a new sperm morphology classification framework that improves model 
performance with no added label costs.
At its core, information is transferred from reliable sperm head masks generated from prior information to an encoder model through unsupervised spatial prediction tasks.
Regularization from masking, label smoothing, and augmentation combine to further facilitate training.
Our three-stage framework was evaluated on the SCIAN and HuSHeM datasets, 
and achieved state-of-the-art results.
Our approach may illuminate a potential for robust, scalable, and accurate 
systems to deal with deteriorating male fertility.

\section{Compliance with ethical standards}
\label{sec:ethics}

This research study was conducted retrospectively using human subject data made available in open access by two publicly available datasets \cite{Shaker2018Hushem,Chang2017GoldstandardFC}. 
Ethical approval was not required as confirmed by the licenses attached with the open access datasets.

\section{Acknowledgements}
\label{sec:experiments}


We would like to acknowledge the professional guidance and support from departments within Anhui Medical University including the First Affiliated Hospital, the NHC Key Laboratory of Study on Abnormal Gametes and Reproductive Tract, the Key Laboratory of Population Health Across Life Cycle, the Anhui Province Key Laboratory of Reproductive Health and Genetics, and the Biopreservation and Artificial Organs Engineering Research Center.

{
    \bibliographystyle{IEEEbib_abbrev}
    \bibliography{refs}
}

\end{document}